\documentclass[runningheads]{llncs}

\usepackage{eccv}

\usepackage{eccvabbrv}

\usepackage{graphicx}
\usepackage{booktabs}

\usepackage[accsupp]{axessibility}  % Improves PDF readability for those with disabilities.

\usepackage[pagebackref,breaklinks,colorlinks,citecolor=eccvblue]{hyperref}
\usepackage{orcidlink}

\begin{document}

% \title{\emph{Ultraman}: Ultra Fast Single Image to Textured 3D Human Reconstruction with Ultra High Resolution} 
\title{Ultraman: Single Image 3D Human Reconstruction with Ultra Speed and Detail}

\titlerunning{Ultraman}

\author{
Mingjin Chen\textsuperscript{*}\inst{1,2} \and
Junhao Chen\textsuperscript{*}\inst{1,3} \and
Xiaojun Ye\inst{1,4} \and
Huan-ang Gao\inst{1} \and \\
Xiaoxue Chen\inst{1} \and
Zhaoxin Fan\inst{5} \and
Hao Zhao\textsuperscript{\textdagger}\inst{1}
}

\authorrunning{Chen et al.}

\institute{Institute for AI Industry Research (AIR), Tsinghua University \\
\email{zhaohao@air.tsinghua.edu.cn} \and
Beijing Normal University - Hong Kong Baptist University United International College \and
Tsinghua Shenzhen International Graduate School, Tsinghua University \and
College of Computer Science, Zhejiang University \and
State Key Laboratory of Multimodal Artificial Intelligence Systems, Institute of Automation, Chinese Academy of Sciences 
}

% \maketitle
\maketitle
\setcounter{footnote}{1}
\begin{center}
    \vspace{-1\baselineskip}
    Project Page: \url{https://air-discover.github.io/Ultraman/} \\
\end{center}\footnotetext{\textsuperscript{*} Indicates Equal Contribution. \textsuperscript{\textdagger} Indicates Corresponding Author.}

\begin{figure}[h]
% \vspace*{-2em}
\centering
\includegraphics[width=0.9\textwidth]{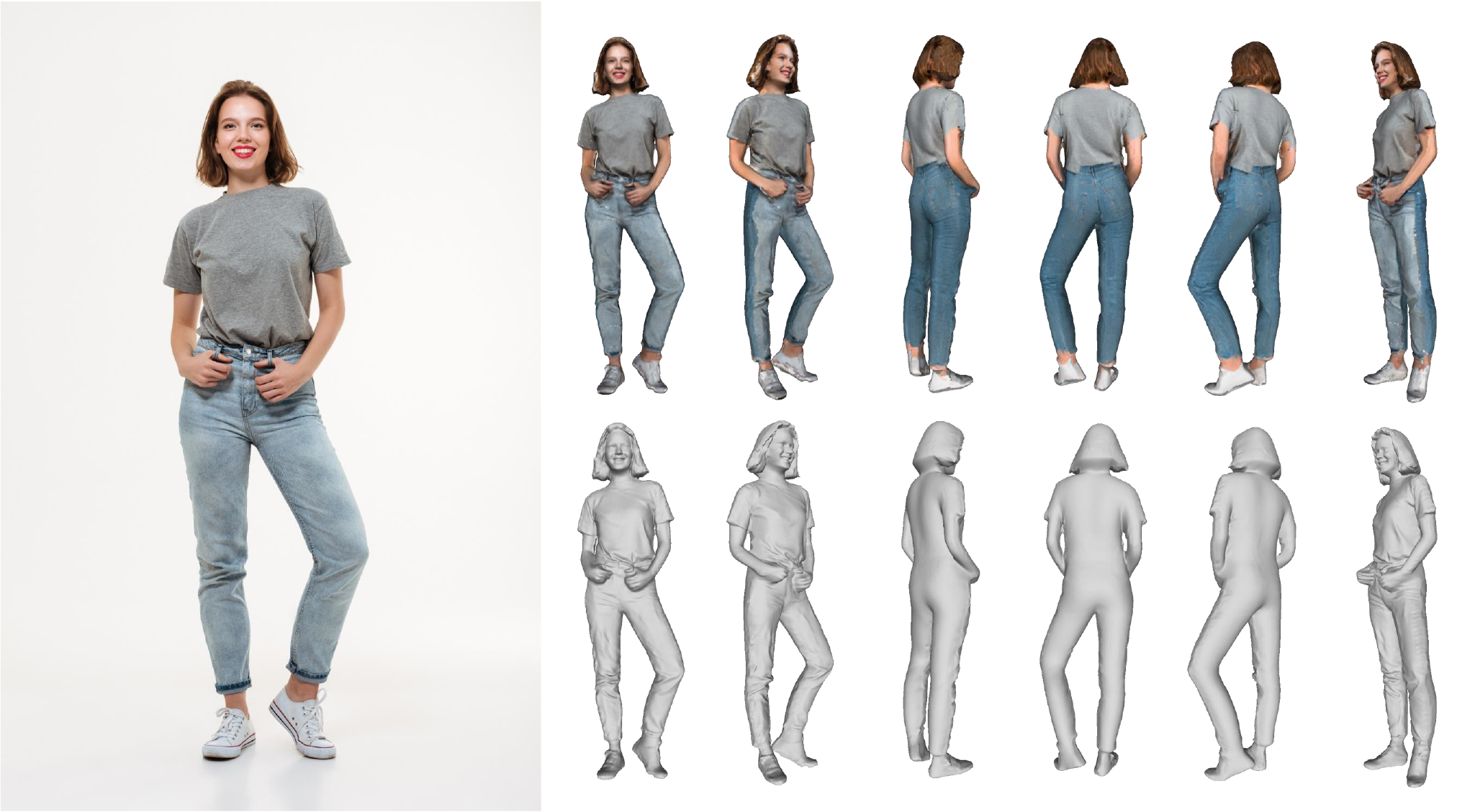}
\caption{Given a single RGB image of a human, \emph{Ultraman} successfully reconstructs the surface geometry and appearance of the clothed human body in 30 minutes, while the state-of-the-art method TeCH \cite{huang2023tech} takes 5 hours on average. It also achieves a high degree of consistency in the back view with the front view, while TeCH struggles to generate meaningful backview appearance.}
\label{fig:teaser}
% \vspace*{-2em}
\end{figure}

\begin{abstract}
3D human body reconstruction has been a challenge in the field of computer vision. Previous methods are often time-consuming and difficult to capture the detailed appearance of the human body.
In this paper, we propose a new method called \emph{Ultraman} for fast reconstruction of textured 3D human models from a single image. Compared to existing techniques, \emph{Ultraman} greatly improves the reconstruction speed and accuracy while preserving high-quality texture details.
We present a set of new frameworks for human reconstruction consisting of three parts, geometric reconstruction, texture generation and texture mapping.
Firstly, a mesh reconstruction framework is used, which accurately extracts 3D human shapes from a single image. At the same time, we propose a method to generate a multi-view consistent image of the human body based on a single image. This is finally combined with a novel texture mapping method to optimize texture details and ensure color consistency during reconstruction.
Through extensive experiments and evaluations, we demonstrate the superior performance of \emph{Ultraman} on various standard datasets. In addition, \emph{Ultraman} outperforms state-of-the-art methods in terms of human rendering quality and speed. Upon acceptance of the article, we will make the code and data publicly available.

\end{abstract}

\section{Introduction}
\label{sec:intro}

3D human reconstruction has been a long-standing problem in both computer vision and graphics fields. Serving as a fundamental component in digital entertainment, virtual reality, and online social networking domains, image-based human reconstruction aims to recover the 3D human shape and texture from images of an individual person. In this paper, we focus on reconstructing the textured 3D human from a single front-view RGB image, which has potential for various applications due to efficient data acquisition and enhanced usability. 

Technically, reconstructing a 3D human from a single image is a challenging problem due to its inherently ill-posed nature. Given that a single image cannot capture the entirety of a human's appearance, it is crucial to infer the geometry and appearance of the invisible parts of the human. This underscores the importance of incorporating 3D priors of humans in the human reconstruction method. Conventional methods \cite{tan2017indirect, zheng2019deephuman, kanazawa2018end} resolve this by introducing parametric human shape models such as SMPL\cite{loper2023smpl} and SCAPE\cite{anguelov2005scape}. However, these methods typically focus solely on reconstructing the human shape without considering appearance. Moreover, due to the limitations of these human models, accurately representing complex and loose clothing worn by humans remains a challenge. Some recent works \cite{han2023high,xiu2023econ,xiu2022icon, huang2023tech} have improved performance in these cases by integrating depth or normal estimation with 3D human reconstruction, leading to more reliable shape estimation results. Despite these advancements, the reconstructed appearance of the human remains lacking in detail or getting unreasonable texture, especially when dealing with high-resolution input images.

% 在这项工作中,我们提出了一个新颖的三维人体重建框架--emph{\emph{Ultraman}},旨在恢复高质量的人体形状和纹理。最初,我们利用基于深度估计的方法从单一模型中提取三维人体形状。随后,我们通过网格简化等后处理技术来增强估计结果。
In this work, we propose \emph{Ultraman}, a novel 3D human reconstruction framework aimed at recovering high-quality shape and texture of the human. We initially utilize a depth estimation-based method \cite{han2023high} to extract the 3D human shapes from a single model. Subsequently, we enhance the estimation results through post-processing techniques such as mesh simplification.

% 为了应对上述在恢复详细人体外观方面的挑战,我们从基于扩散的生成模型的最新进展中汲取了灵感（cite{shi2023mvdream,tang2024mvdiffusion++,liu2023zero,liu2024one}）。

% 我们介绍了一种多视角图像生成模型,它提供了三维人体重建所必需的外观前验。
% 具体来说,该生成模型的输入包括 作为重建参考的原始单视角图像；根据重建形状渲染的深度图像,作为几何先验；提供人体详细描述的增强提示；指定重建所需视点的视点条件。

% 此外,我们还提出了一种视角选择策略,可从包括俯视图和仰视图在内的不同视角覆盖人体,解决了以往方法在头部或足底视角表现不佳的局限性。最后,我们可以使用更新策略将生成的多视角图像逐步投射到归一化三维人体网格的纹理空间中。
To address the aforementioned challenges in recovering detailed human appearance, we draw inspiration from recent advancements in diffusion-based generative models\cite{shi2023mvdream,tang2024mvdiffusion++,liu2023zero,liu2024one, zhang2023adding, ye2023ipadapter}. We introduce a multi-view image generation model that offers the appearance priors essential for 3D human reconstruction. Specifically, the input for this generation model comprises: The original single-view image acting as the reference for reconstruction; A depth image rendered from the reconstructed shape, serving as the geometric prior; An enhanced prompt providing a detailed description of the human; A viewpoint condition specifying the desired viewpoints for reconstruction. In addition, we propose a view selection strategy that covers the human body from various viewpoints, including top and bottom views, which addresses the limitations of previous methods that performed poorly in the head or sole. Finally, we can progressively project the generated multi-view image onto the texture space of the normalized 3D human mesh using an updating strategy.

\cref{fig:teaser} illustrates the reconstruction results achieved with \emph{Ultraman}, showcasing the successful recovery of human gestures and loose clothing with detailed texture. 
% Extensive experiments demonstrate that our method outperforms previous approaches both quantitatively and qualitatively, offering significant benefits for a variety of AR/VR applications. 
A large number of experiments demonstrate that our approach is quantitatively and qualitatively superior to existing methods, laying the foundation for the development of digital humans in AR/VR applications.
To summarize, the contributions of our work include:
\begin{itemize}
    % 我们提出了一种新颖的基于单张图像输入的三维人体重建框架,可实现高质量的人体几何和外观重建。
    \item We present a novel single-image input-based 3D human reconstruction framework for high-quality human geometry and appearance reconstruction.

    % 我们提出了一种基于单张图像作为引导的人体多视角图像生成方法,其输入仅包含一张图像。通过生成对应的提示、深度图像和相机视角,能够准确地生成指定视角的人体图像。
    \item  We present a multi-view image generation method for the human body based on a single image as a guide, whose input contains only one image. By generating corresponding cues, depth images and camera viewpoints, it is possible to accurately generate human body images with specified viewpoints.

    % 我们提出了一种新颖的纹理映射方法,将生成的多视角图像映射回几何上,并且提出了一种优化优化纹理细节的方法,以确保重建过程中的色彩一致性。
    \item We propose a novel texture mapping method to map the generated multi-view images back to geometry and a method to optimize the optimized texture details to ensure color consistency during reconstruction.
    
    % 通过广泛的实验和评估,我们证明了奥特曼在各种标准数据集上的卓越性能。
    \item Through extensive experiments and evaluations, we demonstrate the superior performance of \emph{Ultraman} on various standard datasets. 
    
\end{itemize}

\section{Related Works}

\subsection{3D Human Reconstruction}

3D human reconstruction has been a long-standing problem in computer vision and graphics.  Traditional methods\cite{kanade1997virtualized, starck2007surface,liu2009point} address 3D human reconstruction relying on multi-view images and geometric or photometric constraints. With the rise of deep learning, a series of data-driven methods\cite{tan2017indirect, zheng2019deephuman, kanazawa2018end} emerge and reconstruct humans from images by fitting human shape models such as SMPL\cite{loper2023smpl} and SCAPE\cite{anguelov2005scape}. Besides, some works have been proposed to regress 3D human body directly shapes from images by representing the human
body as 3D meshes \cite{choi2020pose2mesh, lin2021end}, volumetric grids \cite{varol2018bodynet}, or implicit fields \cite{saito2019pifu,saito2020pifuhd} instead of utilizing parametric human models. However, the above methods are typically designed for the human body and lack robustness for loose, clothing topologies. To resolve this, 2K2K\cite{han2023high}, ECON\cite{xiu2023econ} and ICON\cite{xiu2022icon} enhance performance on those cases by incorporating depth or normal estimation with human reconstruction. Moreover, TeCH \cite{huang2023tech} leverages text-to-image diffusion model to optimize the geometry and texture of the clothed humans through multi-view Score Distillation Sampling (SDS). In this paper, we present a novel framework to reconstruct the detailed texture of the 3D human from a single image, which outperforms previous methods on clothed humans.

\subsection{Generative Modeling}
Generative modeling aims to generate novel samples that exhibit similar statistical characteristics to the training samples by learning the underlying distribution. Previous popular generative models include the variational autoencoders (VAEs) \cite{kingma2013auto, rezende2014stochastic, kusner2017grammar, vahdat2020nvae} comprising an encoder network and a decoder network, and Generative Adversarial Networks (GAN) \cite{goodfellow2014generative,brock2018large,qin2020gan,tulyakov2018mocogan,chu2020learning} employing a generator and a discriminator trained competitively. Recently, diffusion models \cite{sohl2015deep} have achieved great success in image synthesis \cite{rombach2022high,zhang2023adding,ye2023ip,podell2023sdxl}, which generate new samples by gradually denoising a normally distributed variable. They also inspired many works to generate 3D assets. Specifically, many works \cite{poole2022dreamfusion, lin2023magic3d,tang2023dreamgaussian,ornek2023syncdreamer} incorporate NeRF\cite{mildenhall2021nerf} or 3D Gaussian\cite{kerbl20233d} with diffusion model and generate 3D objects using text prompt. Besides, a series of works \cite{shi2023mvdream,tang2024mvdiffusion++,liu2023zero,liu2024one} lift 2D diffusion models to generate
consistent multi-view images by introducing viewpoint awareness. In addition, some works \cite{richardson2023texture,chen2023text2tex,youwang2023paint,zeng2023paint3d} focus on generating texture for 3D assets by extending text-conditional generative models. For example, Text2Tex\cite{chen2023text2tex} progressively synthesizes partial textures by incorporating inpainting into a pre-trained depth-aware 2D diffusion model. In this work, we also leverage the text-conditional generative models to enhance the consistency of textures on the human body.

% 1.重建mesh 2.生成贴图 3.贴回mesh
\section{Method}
\begin{figure}[t]
    \centering
    \small
    % \vspace{-0.4cm}
    \includegraphics[width=\textwidth]{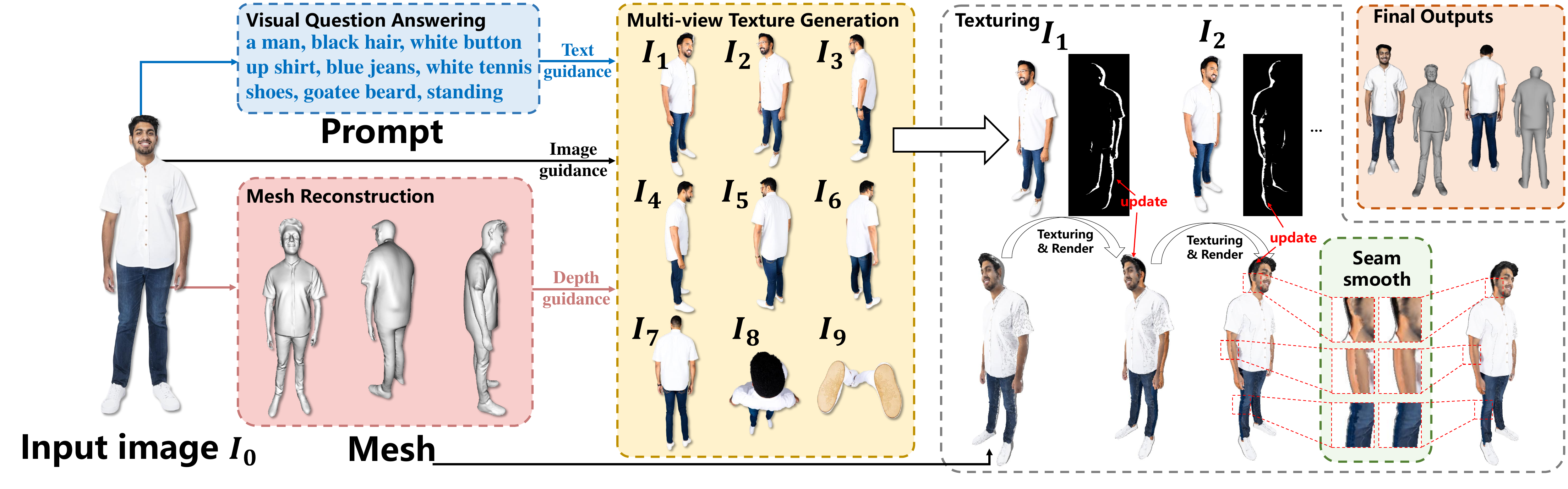}
    % \vspace{-0.3cm}
    \caption{
    Overview of the framework of \emph{Ultraman}. \emph{Ultraman} takes a image $I_0$ of human as input. The blue rounded dashed rectangle is denoted as the Prompt Generation Module and the responses to the questions are generated by GPT4v~\cite{openai2023gpt4v}. The red rounded dashed rectangles are indicated as mesh reconstruction modules. It generates the mesh and UV map. The yellow rounded dashed rectangle is the multi-view texture generation module. It includes a control model $\mathcal{M}_c$, which controls the generation of texture in the current viewpoint by accepting the prompt from the current viewpoint, and by using the depth map rendered by the mesh and the input image. The Texturing module pastes the corresponding texture back onto the mesh according to the generation mask.    }
    % \vspace{-0.5cm}
    \label{fig:mainframework}
\end{figure}

With the goal of 3D human reconstruction from a single image, we introduce a novel framework named \emph{Ultraman}. This framework reconstructs a high-quality body mesh from a single front-view image and completes the invisible parts using a multi-view image generation module and a texturing strategy. It consists of three key modules: the Mesh Reconstruction module, responsible for generating the 3D human mesh and UV maps corresponding to the front view; the Multi-view Image Generation module, which generates images from unobserved views; and the Texturing module, which adds texture to the human body mesh. The overall architecture is illustrated in \cref{fig:mainframework}.

The framework starts by feeding the input image into the Mesh Reconstruction module for mesh reconstruction and UV map export. Concurrently, GPT4v is utilized to respond to questions, facilitating a more detailed description of the person in the input image and enabling the generation of an accurate prompt. Subsequently, the Multi-view Image Generation module accepts the generated prompt and controls the generation of the texture image for the current viewpoint from the depth map in the current viewpoint obtained from the existing mesh rendering and the initial input image. This module contains a redesigned control model consisting of an IP-Adapter~\cite{ye2023ipadapter} and a ControlNet~\cite{zhang2023adding}. In the Texturing module, we use the texture image from the current viewpoint and the corresponding generation mask to add texture to our body mesh. Finally, by determining the gap between the different regions of the generation mask, we smooth the gap to get our output.

% 介绍整个framework的结构
% 印度哥们

\subsection{Mesh Reconstruction} \label{sec:mesh}

Given a high-resolution front-view image $I$ of a human, we aim to reconstruct the 3D human mesh $M$ and corresponding UV maps $U$. To resolve this, we propose a integrated human mesh reconstruction module, which can be formulated as:

\begin{equation}
    M, U = UVG \circ MS \circ MG \circ rembg(I).
\end{equation}

This formulation is composed of four parts, a background removal module $rembg(\cdot)$, a mesh generator $MG(\cdot)$, a mesh simplifier $MS(\cdot)$, and a UV Mesh Generator $UVG(\cdot)$, which are described in details as follows.

\textbf{(1) Background Removal $rembg(\cdot)$}: To reduce the disturbance caused by the background, we preprocess the high-resolution image $I$ with the background removal module\cite{kim2022revisiting}  as $I'=rembg(I)$, enhancing focus on the human body.

\textbf{(2) Mesh Generator $MG(\cdot)$}: The image after background removal $I'$ is used as input to mesh generator $MG$, producing the 3D human mesh with vertex color $M_{vc}=MG(I')$. Specifically, we adopt 2K2K\cite{han2023high} as the baseline and infer 3D human models from single-view images by predicting high-resolution depth maps and merging them into the mesh. 

\textbf{(3) Mesh Simplifier $MS(\cdot)$}: The baseline result $M'_{vc}$ is usually noisy on the surface and has too many vertices which are redundant for the following computation. To address this, we moderately simplify $M'_{vc}$ with quadric edge collapse decimation as $M'_{vc}=MS(M_{vc})$.

\textbf{(4) UV Mesh Generator $UVG(\cdot)$}: With the simplified mesh and vertex color $M'_{vc}$, we generates the Mesh $M$ and its corresponding UV map $U$ through a UV unwrapping tool Xatlas~\cite{xatlas} $UVG(M_{vc})$.

%It is worth noting that the above modules can be replaced with similar baselines, making our formulation a general integrated framework.  

 %With this formulation, we successfully reconstruct the 3D human mesh from single-view images.
 
 This formulation generates high-quality 3D mesh from single-view images and exports them as 3D human mesh and UV maps which correspond to the front face of the human. With the simplification module, the quality of the UV map corresponding to the mesh can be improved, and the texture quality will not be degraded while preserving the existing texture on the front side of the human body. We demonstrated in subsequent experiments that simplifying the mesh can improve the quality of our texture complements. 

\subsection{Consistent Multi-view Image Generation}

\begin{figure}[htbp]
    \centering
    \small
    % \vspace{-0.4cm}
    \includegraphics[width=0.8\textwidth]{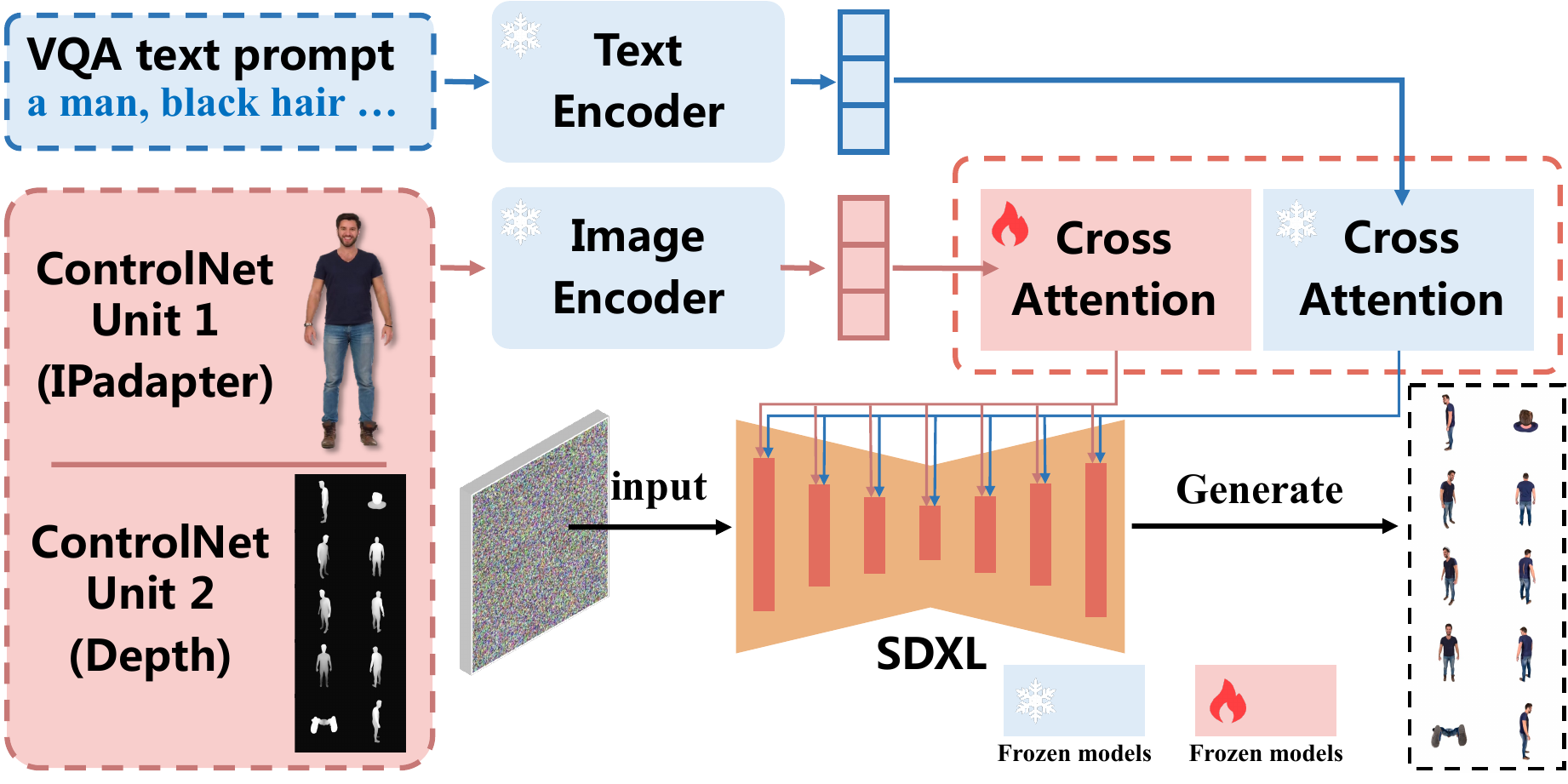}
    % \vspace{-0.3cm}
    \caption{
    We control it by ipadapter and depth as controlnet unit, use a single portrait on the front side as a reference image, and use VQA to generate prompt as a textual cue for SD-XL, and realize the generation of consistent multi-view image on the person. Where the depth map is derived from the depth map rendered from the mesh exported by the Mesh Reconstruction step at the specified viewpoint.
    }
    % \vspace{-0.5cm}
    \label{fig:generationmodel}
\end{figure}

After obtaining the reconstructed 3D human mesh as described in \cref{sec:mesh}, our objective is to produce its texture with the single front view image. Motivated by recent advances in diffusion models, we introduce a multi-view generation model conditioned on depth, views and text to synthesize images of the human from unobserved viewpoints. 
As shown in \cref{fig:generationmodel}.
It is crucial that the generated multi-view image accurately captures the human features depicted in the original single-view image. To achieve this, we utilize a Visual Question Answering (VQA) captioner to provide detailed descriptions of the human's appearance, which serve as prompts for our generation model.

\subsubsection{Prompt Generation} To gather comprehensive details about the character's appearance, we employ the visual language model GPT4V~\cite{openai2023gpt4v} as a Visual Question Answering (VQA) captioner. Rather than using naive image captioning, we formulate a set of specific VQA questions $Q_{vqa}$ covering various aspects such as clothing styles, colors, facial features, and hairstyles. The answers to these questions, denoted as $A_{vqa}$, are then embedded into the prompt $P_{vqa}$ in a specific format. Each prompt $P_i$ consists of both $P_{vqa}$ and location information.

\subsubsection{View Selection Strategy} To generate the complete texture of the human body, we choose 10 views $V = \{v_0, v_1, \ldots, v_9\}$ as input to the multi-view generation model. For each viewpoint, we define $v_i=(\theta, \phi, r) \in V$, where $\theta$ is the azimuth angle with respect to the Z-axis, $\phi$ is the viewpoint elevation angle with respect to the XZ-plane, and $r$ is the distance between the viewpoint and the origin. The viewpoints consist of 8 viewpoints around the character and two additional top/bottom views. The first 8 views in $V$ are horizontal views. We rotate the camera horizontally by angles $\theta$ of {$180^{\circ}$, $0^{\circ}$, $45^{\circ}$, $315^{\circ}$, $90^{\circ}$, $270^{\circ}$, $135^{\circ}$, $225^{\circ}$} relative to the viewpoint of the input image. The remaining 2 views in $V$ represent vertical perspectives. The camera is positioned at elevation angles $\theta$ of {$90^{\circ}$, $270^{\circ}$} to render depth maps directly above and below the human. Simultaneously, we utilize the reconstructed mesh to generate depth maps for each respective view. This view selection strategy ensures comprehensive coverage of the human body from various viewpoints.

%We also generated an image of the back of the human using the control model first. 

%3 

% 3.3.2 脚（底下贴图->解决了裤子内侧的贴图,头顶->解决了）和头贴图生成（这两个关系 ref image）
%\subsubsection{Texture Generation for vertical View}

\subsubsection{Multi-view Image Generation Model} Given the input single-view image, rendered depth, viewpoint and generated prompt, we then build our multi-views generation method on the control model. Specifically, we 
The multi-view image generation formulation is depicted below:
\begin{align}
    I_b &= \mathcal{M}_c\left(I, d_i, P_i, v_i\right), \quad i=0, \\
    I_{i} &=\left\{\begin{array}{ll}
    \mathcal{M}_c\left(I, d_i, P_i, v_i\right), \quad 1 \leq i<6 ,\\
    \mathcal{M}_c\left(I_b, d_i, P_i, v_i\right), \quad 6 \leq i\leq 9.
\end{array}\right.
\end{align}
Here, the control model $\mathcal{M}_c$ contains two pre-trained diffusion models. One is a model IP-Adapter~\cite{ye2023ipadapter} that can generate images from other viewpoints with features consistent with the original single-viewpoint image. Another one is a depth-to-image model Controlnet~\cite{zhang2023adding} that can control the pose of the generated image in other viewpoints with a depth map that has pose consistency with our original input single-view image. $I$ is the front singe-view image of the human, $I_b$ is the back view of the human which is first generated by the control model using the front single-view image,  $d_i$ is the depth map in the $i$-th view, and $P_i$ is the prompt of the corresponding view. It's worth noting that we utilize the back view image $I_b$ as input to the control model for horizontal views positioned behind the character. This approach ensures enhanced consistency of the back view images. Similarly, for the top and bottom views, we also employ $I_b$ as input, which enables more realistic rendering of hair in the top view and more plausible depiction of soles in the bottom view.

% 给一张图
% tech man

% Reconstruction, generation, texturing
\subsection{Texturing}

With the multi-view image generation model, we can produce consistent multi-view images of humans from a single front-view image. In this section, we present the process of projecting the generated multi-view image onto the texture space of the normalized 3D human mesh.
% 与text2tex贴回纹理的方法一致

It's important to note that we do not employ the generation model to synthesize images in the initial viewpoint $v_0$. This is because we aim to preserve the frontal information of the human from our original image. We utilize the generation model to generate images in other viewpoints. For instance, in viewpoint $v_1$, we render the mask $\mathcal{M}^1$, indicating the area requiring texture generation, and then employ the generation model to generate the image $\mathcal{I}_1$. Subsequently, we back-project the image $\mathcal{I}_1$ onto the visible part of the texture $\mathcal{T}^1$. In subsequent steps, we iteratively generate images $\mathcal{I}_i$ in the current viewpoints and back-project them onto the texture $\mathcal{T}^i$.

%与text2tex类似的
\begin{figure}[htbp]
    \centering
    \small
    % \vspace{-0.4cm}
    \includegraphics[width=0.8\textwidth]{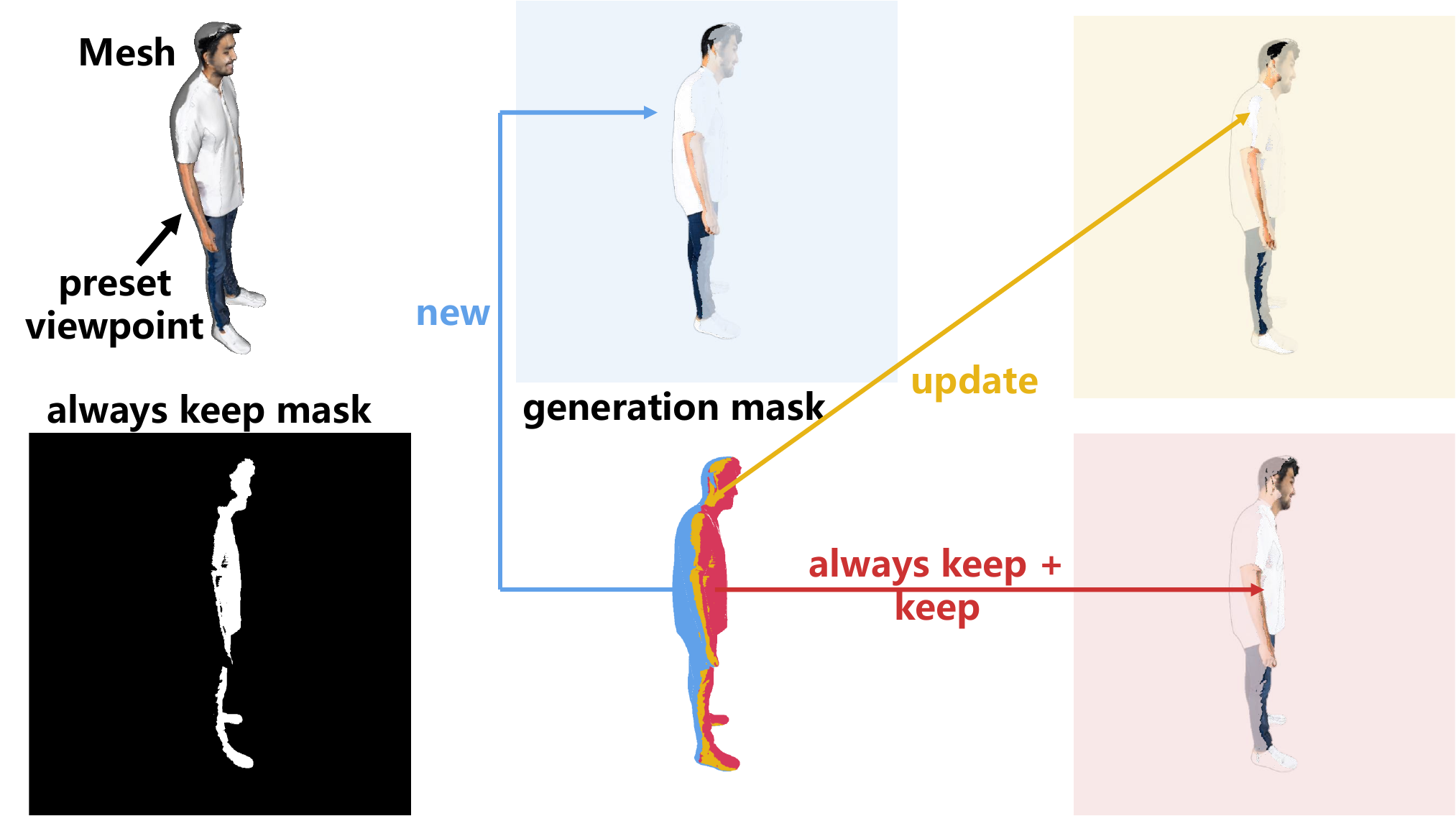}
    % \vspace{-0.3cm}
    \caption{
    We categorize generation masks in the current perspective into four types. We never update the texture in the "always keep" area, which is the front texture of the human body in the initial view and does not need to be regenerated. The "keep" area is the texture that has been best observed in other views and does not need to be updated. The "update" area is where we will update the texture a little bit. The "new" area is where we need to generate new textures.
    }
    % \vspace{-0.5cm}
    \label{fig:texturing}
\end{figure}

\subsubsection{Generation mask}

We adopt a similar strategy to Text2tex \cite{chen2023text2tex}  to determine the regions requiring texture generation. For all viewpoints, we compute the similarity mask $\mathcal{S}^i$ corresponding to each viewpoint $v_i$. With the similarity mask $\mathcal{S}^i$, the mask for the areas needing texture generation $\mathcal{M}^i$ consists of five parts:

\begin{itemize}
   \item  \textbf{Always keep}: This area comprises pixels representing colors already present on the mesh. We retain the pixels in this region to maintain consistency, ensuring that positive features of our textured human mesh remain unchanged.
    \item  \textbf{Keep}: Pixels in this region have been textured and are viewed from a favorable angle according to $\mathcal{S}^i$. Hence, we preserve these pixels.
     \item \textbf{Update}: Pixels in this area have been textured but are better viewed from the current viewpoint according to $\mathcal{S}^i$. We update these pixels based on the generated image.
    \item  \textbf{New}: Pixels in this region are yet to be textured. We texture these areas based on the images we generate.
     \item \textbf{Ignore}: This area contains pixels irrelevant to the body mesh we aim to texture, such as background pixels. Hence, we exclude these pixels from any operations.
\end{itemize}

\begin{figure}[htbp]
    \centering
    \small
    % \vspace{-0.4cm}
    \includegraphics[width=\textwidth]{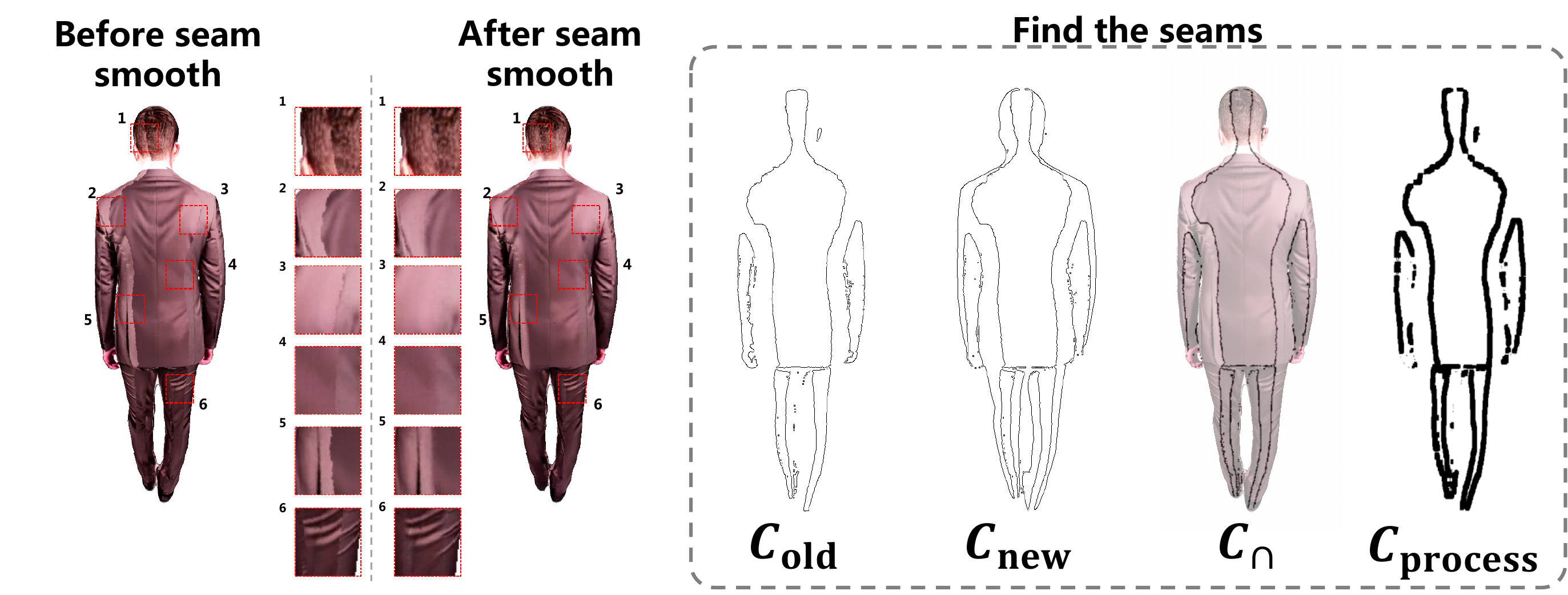}
    % \vspace{-0.3cm}
    \caption{
    Seam smooth effect and process. We show the before and after changes for seam smoothing. We find their canny map through the mask of the old regions that already exist $C_{old}$ and the new regions that are generated in the current viewpoint $C_{new}$. Their canny maps are intersected and extended appropriately. We used the open source toolkit clipdrop for our smooth process. In different viewpoints, $C_{new}$  will be different. When there are no pixels in the current view that need to be updated, we find the canny of the New area mask. When there are no pixels in the current view that need to be updated, we find canny of the mask in the update area.
    }
    % \vspace{-0.5cm}
    \label{fig:seamsmooth}
\end{figure}

\subsubsection{Seam smooth}
After generating an image and texturing it based on the generation mask, seams may appear on the texture. 
As marked by the red box on the left side of \cref{fig:seamsmooth}.
To address this issue, we render an image of the existing texture at this point using the camera parameters from the current viewpoint. We then smooth out the areas where different regions of the generation mask are connected, as these areas may exhibit seams due to variations in the images generated from different angles. Specifically, we obtain the canny map of the masks for the two regions that require seam treatment. The intersection between these two canny maps identifies the areas requiring smoothing. 
$C_{\cap}$ in \cref{fig:seamsmooth} shows the calculated locations of the seams that need to be patched.
The seam smooth function is outlined below:

 \begin{equation}
    I_{s}=\left\{\begin{array}{ll}
    Smooth\left(I_m, C_{keep} \cap C_{new}, v_i \right) & 1 \leq i \leq 4 ,\\
    Smooth\left(I_m, C_{keep} \cap C_{update}, v_i \right) & i=0 \text{ or } 4 <  i\leq 9.
\end{array}\right.
\end{equation}

Here, $I_m$ is the image which rendered from a mesh where a texture already exists. $C_{keep}$ is the canny map of the generation mask of keep region. $C_{new}$ is the canny map of the generation mask of new region. $C_{update}$ is the canny map of the generation mask of update region. $v_i$ is the current viewpoints.

\section{Experiments}

% \subsection{Experimental Setup}
% 
\subsection{Qualitative Comparison} % 定性比较
% 这里放一整页我们的case
\subsubsection{The effect of our method on different cases.}
\begin{figure}[htbp]
    \centering
    \small
    % \vspace{-0.4cm}
    \includegraphics[width=\textwidth]{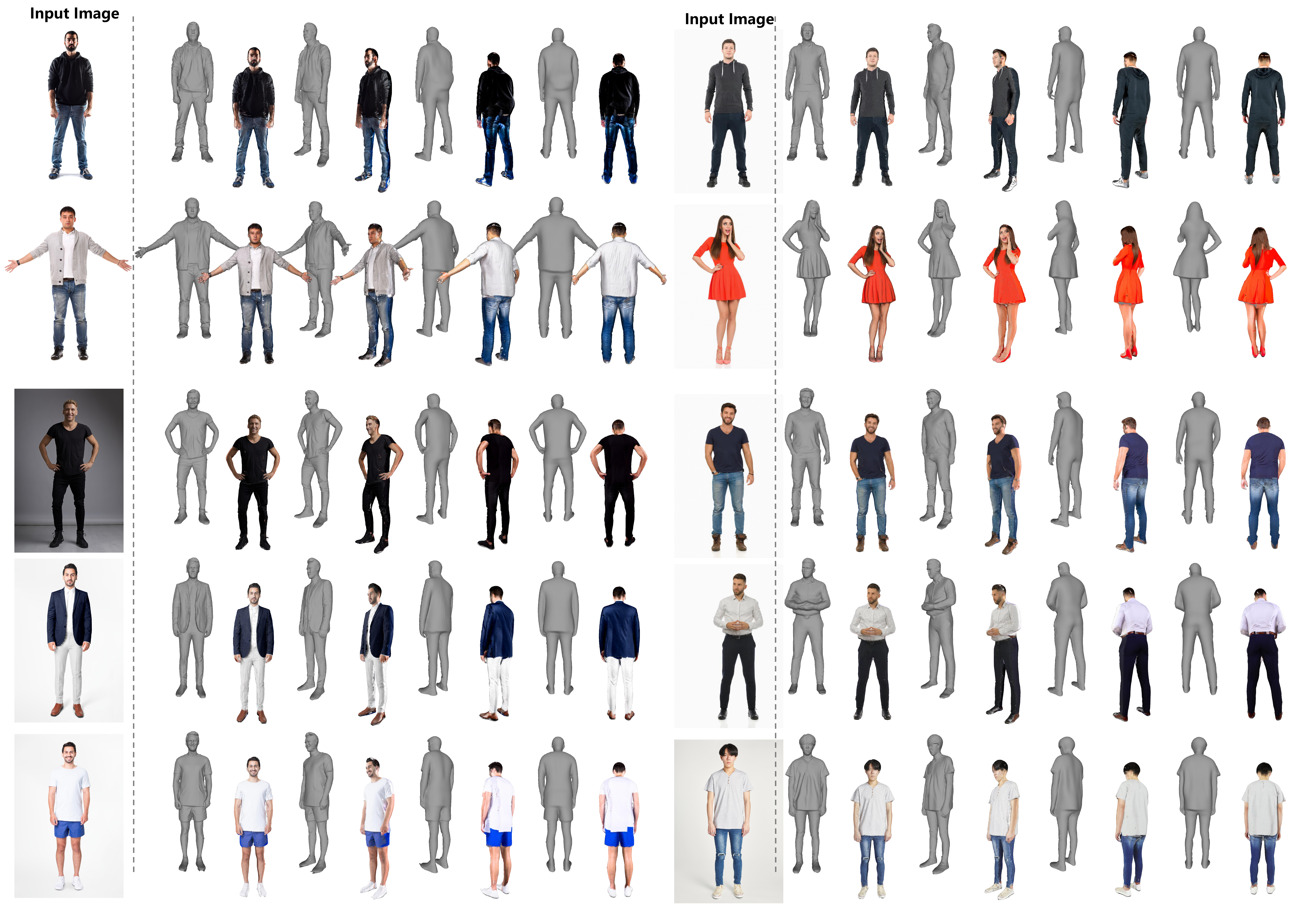}
    % \vspace{-0.3cm}
    \caption{
    \emph{Ultraman} reconstructions on different single portraits.
    }
    % \vspace{-0.5cm}
    \label{fig:morecase}
\end{figure}

The reconstruction results of our \emph{Ultraman} method with different single image inputs are shown in ~\cref{fig:teaser}, \cref{fig:morecase}, \cref{fig:comp}. There are good reconstruction results under different standing postures, different genders, and different dresses. We have a good degree of reproduction in details such as watches, holes in pants, crossed hands or hands in pockets.

\subsubsection{Comparison with existing models.} % 和tech比
\begin{figure}[htbp]
    \centering
    \small
    % \vspace{-0.4cm}
    \includegraphics[width=\textwidth]{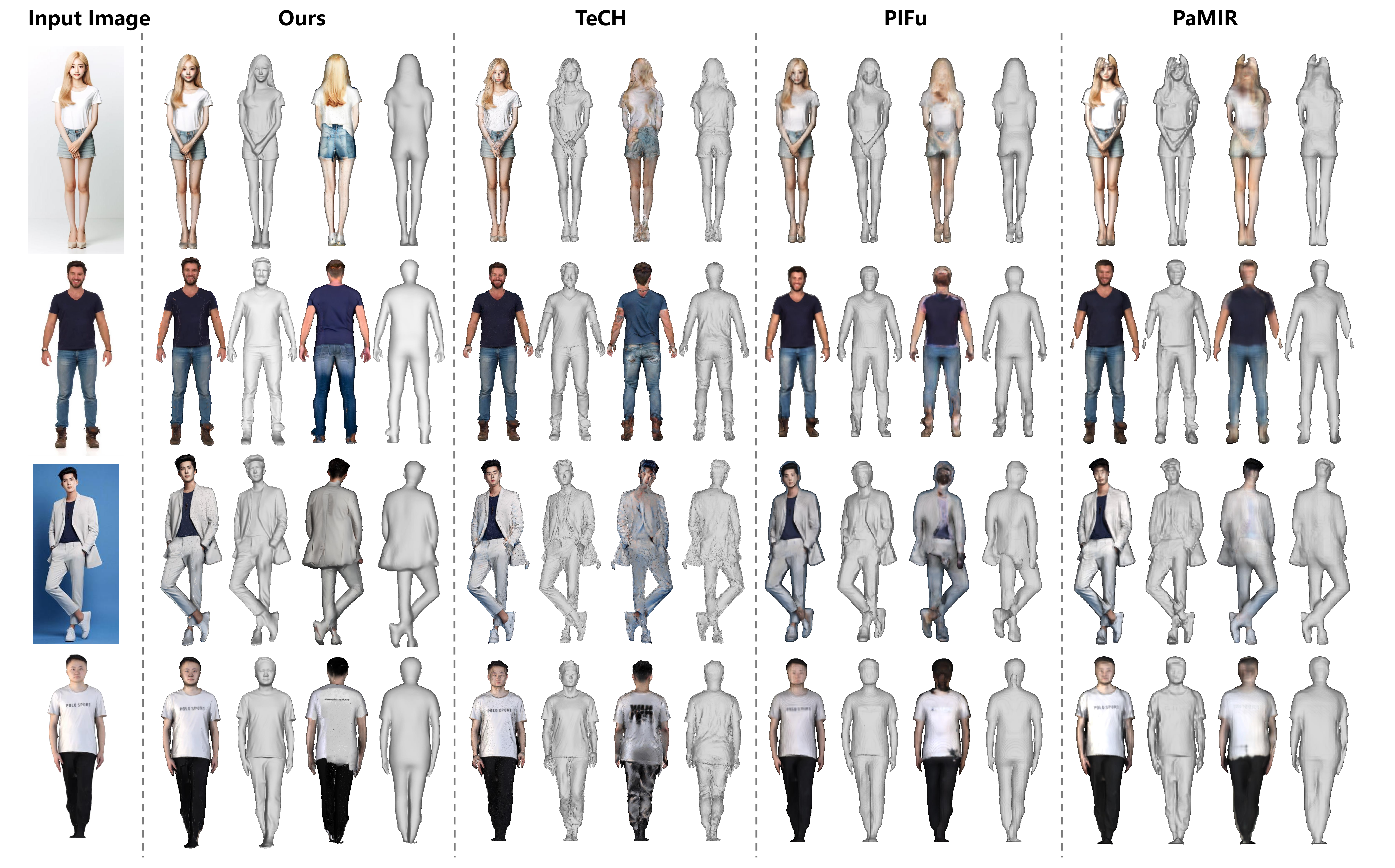}
    % \vspace{-0.3cm}
    \caption{
    Visualization comparison results with TeCH~\cite{huang2023tech}, PIFu~\cite{saito2019pifu}, and PaMIR~\cite{zheng2020pamir}.
    }
    % \vspace{-0.5cm}
    \label{fig:comp}
\end{figure}

We also compared with the current state-of-the-art single-image human reconstruction methods, including TeCH~\cite{huang2023tech}, PIFu~\cite{saito2019pifu}, and PaMIR~\cite{zheng2020pamir}. \cref{fig:comp} shows the visualized comparison results, with \emph{Ultraman} taking the lead in human back mapping generation. And it is also able to clearly distinguish the character features geometrically.
In contrast, TeCH~\cite{huang2023tech}, although having higher quality in geometry, its performance in human back reconstruction is quite poor, showing obvious back texture problems in our experiments and very poor geometric performance on non-fitted clothes.
Both PIFu~\cite{saito2019pifu} and PaMIR~\cite{zheng2020pamir} are far from \emph{Ultraman} in terms of geometry and texture, and \emph{Ultraman} has a much higher quality reconstruction on non-fitted garments.

% \subsubsection{More specific domains.} % 如果这里有一些特殊的case,比如侧身,或者非标bdfgx u7准站姿

\subsection{Quantitative Comparison} % 定量比较
\subsubsection{Inference speed comparison and space} % 时间 and gpu内存占用 简单文字说明一段话即可
\emph{Ultraman} generates a result in about 20-30 minutes. The mesh reconstruction step takes 10 minutes. The generation and texturing steps take about 10 to 20 minutes in total.
For reference, existing SOTA methods typically generate a human mesh with textures in about 4-5 hours, e.g., TeCH. Our method improves 93\% in speed over existing SOTA methods. Our experiments were conducted on the A100. After our tests, our experiments run smoothly and at maximum speed with 32GB of memory. TeCH~\cite{huang2023tech} requires 2*32G GPU memory.

\subsubsection{User study}
We invited users to conduct a preference study. Preference studies were conducted on geometry, texture, and finishing, respectively, while users were asked to choose the best model among those obtained by the 4 methods. The results are shown in \cref{tab:comp}.
The experimental results showed that 90.5\% of the \emph{Ultraman} results were considered by the users as the best results. Although it did not dominate in the geometric comparison with TeCH, it took the lead in texture, especially in the performance of dorsal texture, and in the overall evaluation, only 4.6\% thought that the quality results generated by TeCH were better than that of \emph{Ultraman}. 95.4\% of the users thought that the overall results of \emph{Ultraman} were better than that of TeCH, and more than 98\% of the users thought that the results were better than that of the PIFu with PaMIR.

\subsubsection{Quantitative metrics}
We randomly selected 50 samples for quantitative experiments on Thuman 2.0~\cite{tao2021function4d}. We used four viewpoints of GT, front, back, left and right, and compared the similarity between the reconstructed resultant renderings and GT in these four viewpoints by different methods. CLIP score Image-Image similarity~\cite{radford2021learning}, LPIPS (learned perceptual image path similarity)~\cite{zhang2018unreasonable}, SSIM (Structural Similarity)~\cite{wang2004image}, and PSNR (Peak Signalto-Noise Ratio) scores were tested respectively.
CLIP~\cite{radford2021learning} calculates the similarity between two images, which can better represent the similarity at the level of image features.
LPIPS~\cite{zhang2018unreasonable} metric is widely used to measure the difference between two images based on Learning Perception Image Block Similarity, which is more in line with human perception than the traditional methods.
SSIM~\cite{wang2004image}, compares the similarity between two images in terms of luminance, contrast and structure.
PSNR is based on the error between corresponding pixel points, i.e., based on error-sensitive image quality evaluation.

In the similarity comparison with GT, \emph{Ultraman} takes the lead in all the metrics evaluated. Whether comparing at the pixel level or at the image feature level, \emph{Ultraman}'s results are clearly separated from other methods. As shown in \cref{tab:comp}.

% 定量指标
\begin{table}[htbp]
    \caption{
    Comparisons on CLIP Image-Image similarity, SSIM, LPIPS, PSNR quantitative metrics, and User Study evaluations, with the best results highlighted in bold.
    }
    \label{tab:comp}
    \centering
    \resizebox{\textwidth}{!}{
    \begin{tabular}{l|cccc|cccc}
    \toprule
    Method & \multicolumn{4}{c|}{User Study} & \multicolumn{4}{c}{Quantitative Evaluation} \\
    \cmidrule{2-5} \cmidrule{6-9}
          & \parbox{2cm}{\centering Geometry (Prefer \emph{Ultraman}\%)} & \parbox{2cm}{\centering Texture (Prefer \emph{Ultraman}\%)} & \parbox{2cm}{\centering Overall Satisfaction (Prefer \emph{Ultraman}\%)} & \parbox{2cm}{\centering Which is best (\%)} & \parbox{2cm}{\centering \textbf{CLIP similarity $\uparrow$}} & \textbf{SSIM $\uparrow$} & \textbf{LPIPS $\downarrow$} & \textbf{PSNR $\uparrow$} \\
    \midrule
    \textbf{\emph{Ultraman} (Ours)} & - & - & - & \textbf{90.5\%} & \textbf{0.9131} & \textbf{0.8958} & \textbf{0.1338} & \textbf{17.4877} \\
    TeCH                     & 27.1\% & 98.4\% & 95.4\% & 7.5\% & 0.8875 & 0.8709 & 0.1678 & 15.1464 \\
    PIFu                     & 97.1\% & 97.7\% & 99.9\% &  0.6\% & 0.8501 & 0.8884 & 0.1615 & 15.0248 \\
    PaMIR                    & 92.5\% & 95.4\% & 98.7\% &  1.3\% & 0.8861 & 0.8924 & 0.1431 & 16.6267 \\
    \bottomrule
    \end{tabular}
    }
\end{table}

% \begin{table}[htbp]
%     \caption{Performance evaluation of different model configurations under the Ours (\emph{Ultraman}) framework in the user study, including both user evaluations and quantitative assessments.}
%     \label{tab:ours_method_configs_extended}
%     \centering
%     \resizebox{\textwidth}{!}{ % Adjusted to fit the added columns
%     \begin{tabular}{c|c|ccc|cccc}
%     \toprule
%     Reconstruction & Generation & \multicolumn{3}{c|}{User Study} & \multicolumn{4}{c}{Quantitative Evaluation} \\
%         \cmidrule{3-5} \cmidrule{6-9}  % Add lines under the group titles
%     Model          & Model      & \parbox{2cm}{\centering Geometry (\%)} & \parbox{2cm}{\centering Texture (\%)} & \parbox{2cm}{\centering Overall Satisfaction (\%)} & \parbox{2cm}{\centering \textbf{CLIP similarity $\uparrow$}} & \textbf{SSIM $\uparrow$} & \textbf{LPIPS $\downarrow$} & \textbf{PSNR $\uparrow$} \\
%     \midrule
%     2k2k & SD-XL+IPadapter+depth & 78 & 74 & 76 & \textbf{0.9131} & \textbf{0.8958} & \textbf{0.1338} & \textbf{17.4877} \\
%     2k2k & SD-XL+depth          & 81 & 77 & 79 & 0.88 & 0.96 & 0.18 & 26.1 \\
%     ECON & SD-XL+IPadapter+depth & 75 & 71 & 73 & 0.82 & 0.93 & 0.22 & 24.6 \\
%     ECON & SD-XL+depth          & 78 & 74 & 76 & 0.85 & 0.94 & 0.21 & 25.0 \\
%     ICON & SD-XL+IPadapter+depth & 82 & 78 & 80 & 0.90 & 0.97 & 0.17 & 26.8 \\
%     ICON & SD-XL+depth          & 85 & 81 & 83 & 0.92 & 0.98 & 0.15 & 27.4 \\
%     \bottomrule
%     \end{tabular}
%     }
% \end{table}

\subsection{Ablation Study}
We selected a portion of samples from the Thuman 2.0 dataset and some in-the-wilds images for our ablation experiments.
\subsubsection{Does Seam smooth help?}
We show in \cref{fig:seamsmooth}, seam smooth is essential for generating smoother and more comfortable textures. In \cref{fig:ablation}, it is also clear that without the Seam smooth effect, the generated texture does not pass through the gaps that can be clearly seen. The result is an uncomfortable look and feel. The various metrics shown in the \cref{tab:ablation} have also declined due to the loss of efficacy of Seam smooth.

\subsubsection{Does VQA help?}
The result presented in the \cref{fig:ablation} demonstrate that VQA notably enhances color consistency between texture of the back and the front input image. This improvement is further supported by \cref{tab:ablation}, across all quality metrics.

\subsubsection{Does IPadapter help?}
\begin{figure}[htbp]
    \centering
    \small
    % \vspace{-0.4cm}
    \includegraphics[width=0.8\textwidth]{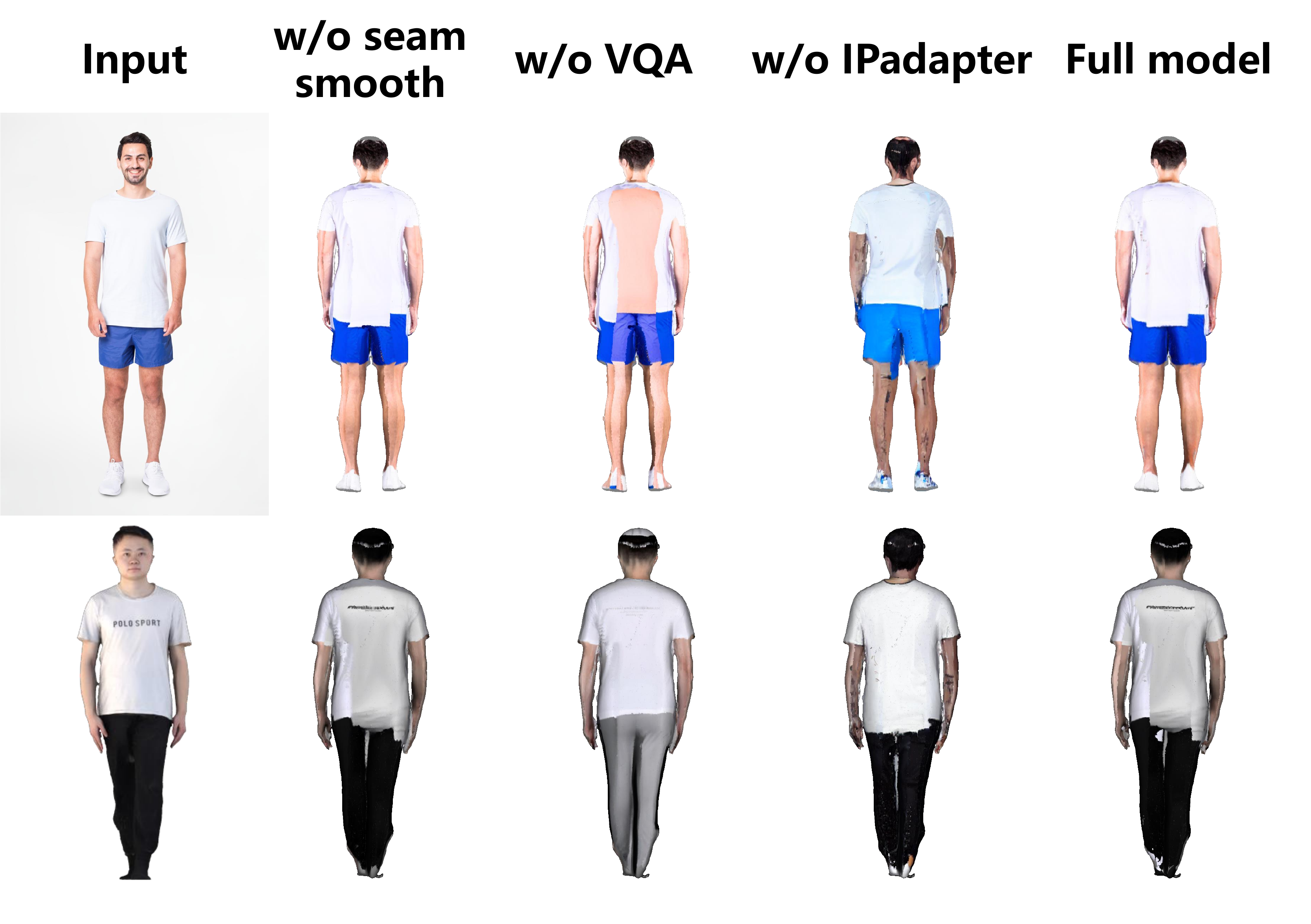}
    % \vspace{-0.3cm}
    \caption{
    \textbf{Ablation study.} To assess the effectiveness of the individual components. We quantitatively eliminate the efficacy of each component. Applying seam smoothing, VQA and IPadapter scheme effectively improves the quality of human synthetic textures.
    }
    % \vspace{-0.5cm}
    \label{fig:ablation}
\end{figure}

As shown in \cref{fig:ablation}, it can be visually noticed that the color of the texture on the back is more vivid and realistic after using the IPadapter and the generated back texture is more consistent with the input front image. This significant improvement is also reflected by the quality metrics in the \cref{tab:ablation}.

% 消融实验的图

% 消融实验表格

\begin{table}[htbp]
\centering
\caption{Ablation study for \emph{Ultraman} components.}
\resizebox{0.7\columnwidth}{!}{
\begin{tabular}{lcccc}
\toprule 
\textbf{Model} & \textbf{CLIP similarity $\uparrow$} & \textbf{SSIM $\uparrow$} & \textbf{LPIPS $\downarrow$} & \textbf{PSNR $\uparrow$} \\ \hline
w/o VQA       & 0.8907 & 0.8889 & 0.1374 & 17.2321 \\ 
w/o seam smooth & 0.9040 & 0.8869 & 0.1347 & 16.8192 \\ 
w/o IPadapter & 0.8705 & 0.8925 & 0.1393 & 17.4040 \\ 
full model    & 0.9131 & 0.8958 & 0.1338 & 17.4877 \\
\bottomrule
\end{tabular}
}
\label{tab:ablation}
\end{table}

\section{Conclusion}

 This paper presents a new 3D human reconstruction framework called \emph{Ultraman} for fast reconstruction of textured 3D face models from a single image. This framework comprises three modules, a mesh reconstruction module, a multi-view image generation model, and a texturing model. Specifically,  we first present a mesh reconstruction module that accurately extracts 3D human shapes from a single image. Then, we introduce a multi-view image generation model and incorporate it with a novel texture mapping method to optimize texture details and color consistency during reconstruction. Compared with previous arts, \emph{Ultraman} improves the reconstruction speed and accuracy while ensuring high-quality texture details. Extensive experiments demonstrate the superior performance of \emph{Ultraman} and show potential in various downstream applications including digital entertainment, virtual reality, and online social networking.
 
% \subsubsection{Limitations} 

\clearpage  % TODO REVIEW/FINAL: This \clearpage needs to be removed from both review and camera-ready versions.

\bibliographystyle{splncs04}
\bibliography{main}
\end{document}